\documentclass[]{article}
\usepackage[letterpaper]{geometry}
\usepackage{mtsummit2021}
\usepackage{times}
\usepackage{url}
\usepackage{latexsym}
\usepackage{natbib}
\usepackage{layout}

\usepackage{caption}
\usepackage{subcaption}
\usepackage{adjustbox}

\newcommand\blfootnote[1]{%
  \begingroup
  \renewcommand\thefootnote{}\footnote{#1}%
  \addtocounter{footnote}{-1}%
  \endgroup
}

\begin{document}
	
	\blfootnote{Accepted for the  1st International Workshop on Automatic Translation for Signed and Spoken Languages (ATS4SSL), August 20, 2021}

	\title{\bf Sign and Search: Sign Search Functionality for Sign Language Lexica}
	
	\author{\name{\bf Manolis Fragkiadakis} \hfill  \addr{m.fragkiadakis@hum.leidenuniv.nl}\\
		
		\addr{Leiden University Centre for Digital Humanities, Leiden University,
			Leiden, 2311VJ, the Netherlands}
		\AND
		\name{\bf Peter van der Putten} \hfill \addr{p.w.h.van.der.putten@liacs.leidenuniv.nl}\\
		\addr{Leiden Institute of Advanced Computer Sciences, Leiden University,
			Leiden, 2333CA, the Netherlands}
	}

	\maketitle
	\pagestyle{empty}
	
	\begin{abstract}
		Sign language lexica are a useful resource for researchers and people learning sign languages. Current implementations allow a user to search a sign either by its gloss or by selecting its primary features such as handshape and location. This study focuses on exploring a reverse search functionality where a user can sign a query sign in front of a webcam and retrieve a set of matching signs. By extracting different body joints combinations (upper body, dominant hand's arm and wrist) using the pose estimation framework OpenPose, we compare four techniques (PCA, UMAP, DTW and Euclidean distance) as distance metrics between 20 query signs, each performed by eight participants on a 1200 sign lexicon. The results show that UMAP and DTW can predict a matching sign with an 80\% and 71\% accuracy respectively at the top-20 retrieved signs using the movement of the dominant hand arm. Using DTW and adding more sign instances from other participants in the lexicon, the accuracy can be raised to 90\% at the top-10 ranking. Our results suggest that our methodology can be used with no training in any sign language lexicon regardless of its size.
	\end{abstract}

	\section{Introduction}
	Sign language lexica are a valuable source for people learning sign languages, teachers and parents who need to communicate in signs with their deaf children as well as researchers studying the languages in question. These lexica allow the user to submit a query containing a unique identifier that by definition refers to a sign (commonly referred to as gloss) and retrieve a video or an image of that sign. In addition to this functionality, some lexica let the user define the formal parameters of the target sign (i.e. its location, handshape, or movement) and retrieve all the signs that contain these features. It is then at the users' discretion to view all the provided signs and select the desired ones. These search functionalities are particularly useful as sign languages, contrary to spoken languages, do not have any unified notation system for sign representation.
	
	Even though a sign search functionality which is based on formal parameters is a user-friendly option in sign language lexica, it still requires manual annotation. Dictionary compilers have to manually link these values to the individual videos of signs. This is a time consuming and prone to errors task and, as \cite{zwitserlood_sign_2010} discusses, it is the reason why only a few of such dictionaries exist to date. More importantly, according to Zwitserlood, these dictionaries are unidirectional ``giving only signed translations of words from a spoken language in a one-to-one relation'' \citep{zwitserlood_sign_2010}. Furthermore, as the retrieved results contain only the parameters selected by the user, the signs are presented in no particular order. 
	
	In this paper, we describe a methodology and its experimental results for multi-directional search functionality for sign language lexica. Our proposed method, extending on previous efforts by \cite{schneider_gesture_2019} and \cite{fragkiadakis_signing_2020}, utilizes either the Uniform Manifold Approximation and Projection for Dimension Reduction (UMAP) technique or the Dynamic Time Warping (DTW) algorithm to measure the distance between a query sign and all the signs in a lexicon. Both techniques require no training, thus making our methodology applicable to any sign language. Requiring training would add further obstacles to making dictionaries widely available, similar to how the need for manual annotation is limiting dictionary availability. Finally, these methods have been compared to two other techniques namely Principal Component Analysis (PCA) and Euclidean distance.  
	
	The paper is organized as follows: in Section 2 we discuss the research which has been conducted in relation to search functionality for sign language lexica or has the potential to be applied in that domain. In Section 3 we describe our methodology regarding the extraction of the body joint coordinates as well as the methods and algorithms compared in this study. In Section 4 we present the results of our experiments. We discuss them in Section 5 and conclude and motivate future research in Section 6.

	\section{Related Work}
	\label{related}
	
	Over the last decade, many research projects have examined the use of computer-vision techniques to allow a user to search a sign in a database or lexicon by performing it in front of a camera or sensor. \cite{hutchison_system_2012} have developed a system for  semi-automatic search functionality. In their system, a user marks the start and end frames of a sign and denotes whether the sign is one- or two-handed. Consequently, the system detects the hands on the basis of skin color and motion. The user can correct, if needed, the detected hand locations and pass the query to the system. Using Dynamic Time Warping their approach computes the similarity between the query sign and all the signs in the database. Their results suggest a 78\% accuracy on the top-10 retrieved signs on a 1113 sign lexicon. While the accuracy rate is high enough, the user still needs to indicate the handedness feature (one- or two-handed) as well as the duration of the sign. Additionally, the data-set used in this study has been recorded under studio conditions posing the question of applicability on noisy real-life conditions on the video query.
	
	\cite{conly_integrated_2015} have used Dynamic Time Warping to match a sign on an American Sign Language dictionary. Using Microsoft's Kinect they detect the hand positions and perform sign matching. Their results suggest an accuracy of 77.3\%  on the top-50 retrieved signs. A major advantage over Wang's et al. (\citeyear{hutchison_system_2012}) implementation is that this system does not require the intervention of the user.
	
	\cite{metaxas_linguistically-driven_2018} have developed a framework that analyzes handshape, orientation, location, and motion trajectories to recognize 350 ASL signs. By passing the extracted features into Hidden Conditional Ordinal Random Fields (HCORF) they achieve a top-1 accuracy of 93.3\% and a top-5 accuracy of 97.9\%. 
	
	\cite{yauri_vidalon_brazilian_2016} have created a system for Brazilian Sign Language recognition using Dynamic Time Warping, a Nearest-Neighbor classifier and Kinect. On a data-set of 107 signs, they have reported an accuracy of approximately 98\%. A major drawback of their results is the fact that their data-set is user-dependent. 
	
	The majority of the aforementioned studies use either a depth sensor or computer-vision techniques. These techniques primarily rely on color and motion detection algorithms, as feature extraction methods, which imposes additional problems. Such techniques can be prone to errors and most importantly require studio conditions in order to predict the required features such as the face and the hands. While such conditions can be true for the videos in a lexicon they cannot be assumed for the query videos. Searching a lexicon can be done in any possible space and lighting conditions, thus it is important that the technique used to capture the required features to be as much as inclusive as possible.
	
	In 2017 \cite{cao_realtime_2017} presented a framework for multi-person 2D pose estimation, OpenPose. This framework can efficiently detect body, foot, hand and facial key-points from a simple RGB video or picture. Its high accuracy, performance and easy implementation make it the ideal framework to parse sign language and gestural videos. Its output consists of multiple json (or differently formatted) files containing all the pixel x, y coordinates of the body, hand and face joints. Most studies use OpenPose to pre-process the videos and use its output to further train or compare machine and deep learning models.
	
	\cite{schneider_gesture_2019} have used OpenPose as well as DTW and Nearest-Neighbor algorithm to perform classification of six gestures. Their results suggested an accuracy of 77.4\%. Most recently, \cite{fragkiadakis_signing_2020} have used OpenPose and DTW to predict a sign recorded using a webcam from a 100 signs lexicon. Their method predicted the matching sign with an 87\% and 74\% accuracy at the top-10 and top-5 retrieved signs by using the path of the dominant hand's wrist.  
	
	This study extends on previous efforts for efficient sign ranking for sign language lexica by: 
	
	\begin{itemize}
		\item Considering a far larger lexicon compared to previous efforts: 1200 signs in total
		\item Comparing four different techniques: Principal Component Analysis (PCA), Uniform Manifold Approximation and Projection for Dimension Reduction (UMAP),  Dynamic Time Warping (DTW) and Euclidean distance
		\item Comparing three different skeletal joint combinations (upper body, dominant hand arm, dominant hand wrist)
		\item Exploring potential accuracy increase by adding more sign instances in the lexicon
	\end{itemize}

	An important difference from previous studies in search functionality for sign language dictionaries is that in our case we expect signers to not “properly" sign a particular sign. As \cite{alonzo_effect_2019} discuss, it is possible that people would not remember exactly how a sign is performed, and as a result, they might sign it slightly differently. Thus it is expected that the matching sign would not be in the first retrieved sign result. This is precisely the reason why we tested our methodologies on a data-set that contains signs performed also by people with no or little experience in sign language. In most sign language data-sets used for sign language recognition tasks, signs are mostly performed by people familiar with sign languages. However, sign language lexica are intended also for people with little knowledge of sign language. As a result, high variability is expected when recording a sign.
	
	Another limitation posed in our study is that sign language lexica do not often contain multiple instances of a particular sign. While various studies using deep learning techniques have shown high accuracy in predicting different signs \citep{li2020word, gokcce2020score, sincan2020autsl, hosain2021hand}, they cannot be used in our case. These techniques often require vast amount of data in order to be trained which might not be available on all sign language lexica. Our main goal is to develop a system that can be easily used in any sign language lexicon regardless of the amount of data in it and most importantly the language itself. However, in this study, we explore the possibility of having a few additional sign instances in the lexicon and their potential benefit to successful sign retrieval.

	\section{Data-sets and Methods}
	In this section we describe the data pre-processing as well as the data-sets and methods used in this study.
	
	\subsection{Data Pre-processing and Normalization}
	OpenPose outputs x, y pixel coordinates for each predicted body and finger joint. These pixel coordinates are relative to the frame size and as a result it is important to normalize them to account for different positions in the frame. As all people in the data-set (both in participants' and lexicon's data) are expected to be in an upright position in front of the camera, rotational in-variance is omitted. The normalization process is the following: for each detected person in a frame, the neck key-point coordinates are subtracted from all the other key-points. Subsequently, all key point coordinates are being divided by the distance between the left and right shoulder key-point. Finally, a horizontal flip is applied when a participant is left-handed by calculating the average velocity of each hand's wrist. The overall normalization process is based on previous studies by  \cite{celebi_gesture_2013}, \cite{schneider_gesture_2019} and \cite{fragkiadakis_signing_2020}. 
	Furthermore, all signs have been re-interpolated to 86 frames which is the mean sign length. Although it makes little difference to DTW’s accuracy, equal length inputs make it easier to handle.
	
	\subsection{Data-sets}
	For this study we used the Ghanaian Sign Language lexicon (GSL)  \citep{fragkiadakis_ghanaian_2021, handslab_ghanaian_nodate}. This lexicon consists of 1200 signs from one signer and has been compiled for educational purposes to be used in a mobile application. A lot of studies in the sign language recognition field have used sign language data-sets from well documented sign languages with primarily signers with light skin tones. We have decided to apply our methodology in a sign language less documented and analyzed with computer vision and machine learning algorithms in order to further explore how these techniques can perform in such conditions.
	
	In addition, the data gathered by \cite{fragkiadakis_signing_2020} have been used to compare the different algorithms described in the next section. This data-set contains the data of ten participants. Each one of them performed the same 20 signs, from the original lexicon, in front of a webcam. The data of two participants have been discarded due to inconsistencies of OpenPose on recognizing their right-hand finger’s and left arm joints.
	
	We have decided to include in the lexicon the data from a random participant every time we tested the methodology. As the lighting conditions on the participants' videos were of poor quality, the predicted body joints by OpenPose had substantially more noise compared to the ones predicted on the lexicon's data. By extending the database with another participant's data, we introduced some noise to the otherwise non-noisy data-set. As a result, each participant’s sign was compared with 1220 signs in our database (1200 from the GSL lexicon and 20 from another random participant). A complete overview of the participants' data-set and the apparatus used to gather the data can be found in Fragkiadakis's et al. (2020) study.
	
	
	One of the main goals of this study is to find if and how different skeletal joints affect the accuracy of the algorithms. As a result, we have compiled 3 different data-sets per condition per participant’s data. The first data-set contains the upper body joints as well as the dominant hand fingers joints’ coordinates resulting in a $86\times29\times2$ (frames  \emph{by} skeletal joints \emph{by} x, y coordinates) dimensionality per sign. Consecutively, the second data-set contains the dominant hand arm joints’ coordinates (nose, neck, shoulder, elbow, wrist) resulting in a $86\times5\times2$ dimensionality per sign. Finally, the data-set regarding the dominant hand wrist data has a $86\times2$ dimensionality per sign.

	\subsection{Methods}
	The following section describes the methods and the four techniques used in this study.
	
	\subsubsection{Dimensionality Reduction}
	As described in the previous section, each sign in each compiled data-set can be seen as a multidimensional vector. To properly project it into the 2D space while still retaining most of the original information, we used two dimensionality reduction techniques. 
	
	The first technique applied is Principal Component Analysis (PCA). PCA is an orthogonal linear transformation that converts the data to a new frame of reference. PCA constructs Principal Components as linear combinations of the initial variables. These components are not correlated and most of the information within the introductory variables is compressed into the first components. By disposing the components with low information and taking into account the remaining ones as new variables, it allows for dimensionality reduction without loosing information. As a technique it has been widely used in the gestural as well as the sign language domain either as a visualization technique or as a pre-processing stage prior to other machine and deep learning stages \citep{gweth_enhanced_2012,sawant_real_2014,haque_two-handed_2019,gao_chinese_2021}. 
	
	Furthermore, the Uniform Manifold Approximation and Projection for Dimension Reduction (UMAP) technique has been utilized. This method has been used instead of another popular dimensionality reduction technique called t-distributed stochastic neighbor embedding (T-sne) \citep{van_der_maaten_visualizing_2008}.
	T-sne's inability to preserve the global structure of the data makes it unusable if distances between different clusters or points need to be calculated such as in our case \citep{mcinnes_umap_2020}. In contrast, UMAP can better preserve both local and most of the global structure in the data allowing the calculation of distance metrics between clusters. Moreover, the lack of normalization in UMAP effectively reduces the time of computation of the high-dimensional graph.
	
	In our study both PCA and UMAP have been used to reduce the dimensionality of each sign to a single x, y coordinate. Subsequently, we measured the euclidean distance between all the signs of the lexicon and the participants' signs. Accuracy for each participant's sign was measured based on whether the target sign was on the top-k shortest distant signs. 
	
	Furthermore, in order to validate the results produced by the UMAP algorithm in its ability to preserve the global distances of the data, we calculated their euclidean distances in the original high-dimensional space. This method has been used as a benchmark to compare the results of both PCA and UMAP.
	
	\subsubsection{Dynamic Time warping}
	
	In addition to the dimensionality reduction techniques described above, Dynamic Time Warping (DTW) has been used to measure the similarity between the different signs.
	Dynamic Time Warping is a dynamic programming based time series comparison algorithm to produce a distance metric between two inputs. It has been widely used in the speech recognition domain \citep{myers_performance_1980, abdulla_cross-words_2003,axelrod_combination_2004} as well as the gestural and sign language recognition fields as shown in Section \ref{related}.
	
	In this study we utilize a DTW python implementation with open beginning and ending attributes by \cite{giorgino_computing_2009} and \cite{tormene_matching_2009} which in a preliminary experiment produced better results compared to the previous DTW implementation by \cite{fragkiadakis_signing_2020}. 
	Similarly, we used a median filter with radius $r = 3$ for smoothing the time series signals from the body joints.
	
\begin{figure}[!h]
		\centering
		\includegraphics[width=0.5\linewidth]{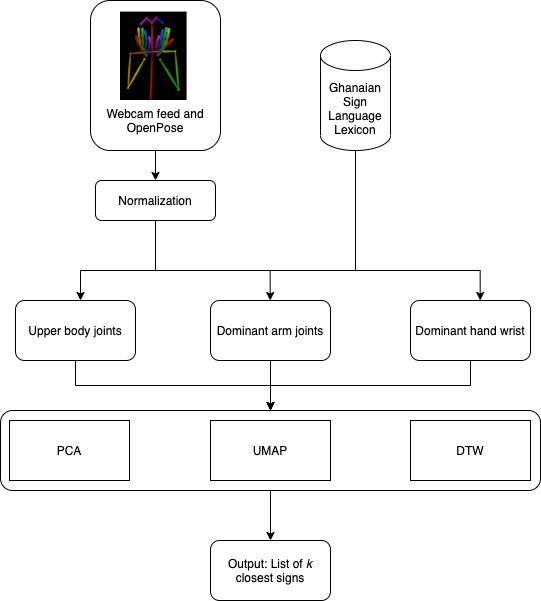}
		\caption{Pipeline of the overall study.}
		\label{fig:Pipeline_1}
	\end{figure}

	Finally, the overall pipeline of the experiment can be seen in Figure \ref{fig:Pipeline_1}.
	
	\subsubsection{How many signs?}
	Many sign language lexica allow their users to submit their own versions of signs. As a result, different instances of the same sign can be stored in the database. One of our research questions is whether having multiple instances of each sign can potentially improve the accuracy of the algorithms. To verify that, we progressively added the 20 signs from other participants to the lexicon. Subsequently, we measured the average top-1 and top-10 accuracy for each algorithm and each skeletal condition. 
	
	Such information can be useful to sign language lexicographers when compiling sign language lexica. They can take advantage of crowd-sourcing material, contributing not only to the augmentation of their lexica but also to the accuracy of the models used for enhanced search functionality.

	\section{Results}
	Table 1 presents the overall accuracy for each of the skeletal conditions. Top-k refers to the number of signs a user must look up before finding a correct match. Accuracy indicates whether the target sign is present in the top-k retrieved signs and is averaged across all participants and all signs. 
	
\begin{figure*}
		\centering
		\begin{subfigure}[b]{0.3\textwidth}
			\centering
			\includegraphics[width=0.85\linewidth]{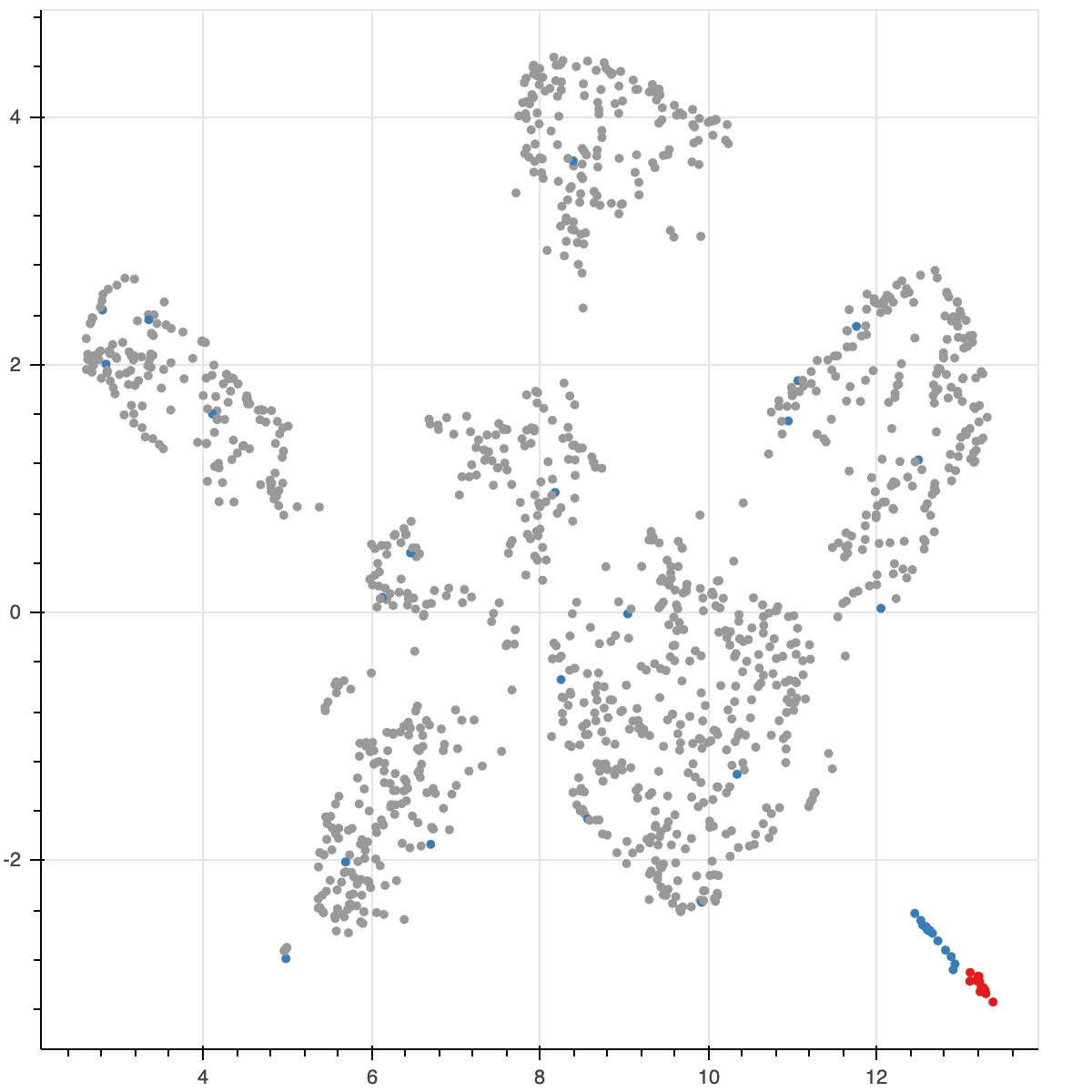}
			\caption{upper body}
			\label{fig:pca_upper_body}
		\end{subfigure}
		\hfill
		\begin{subfigure}[b]{0.3\textwidth}
			\centering
			\includegraphics[width=0.85\linewidth]{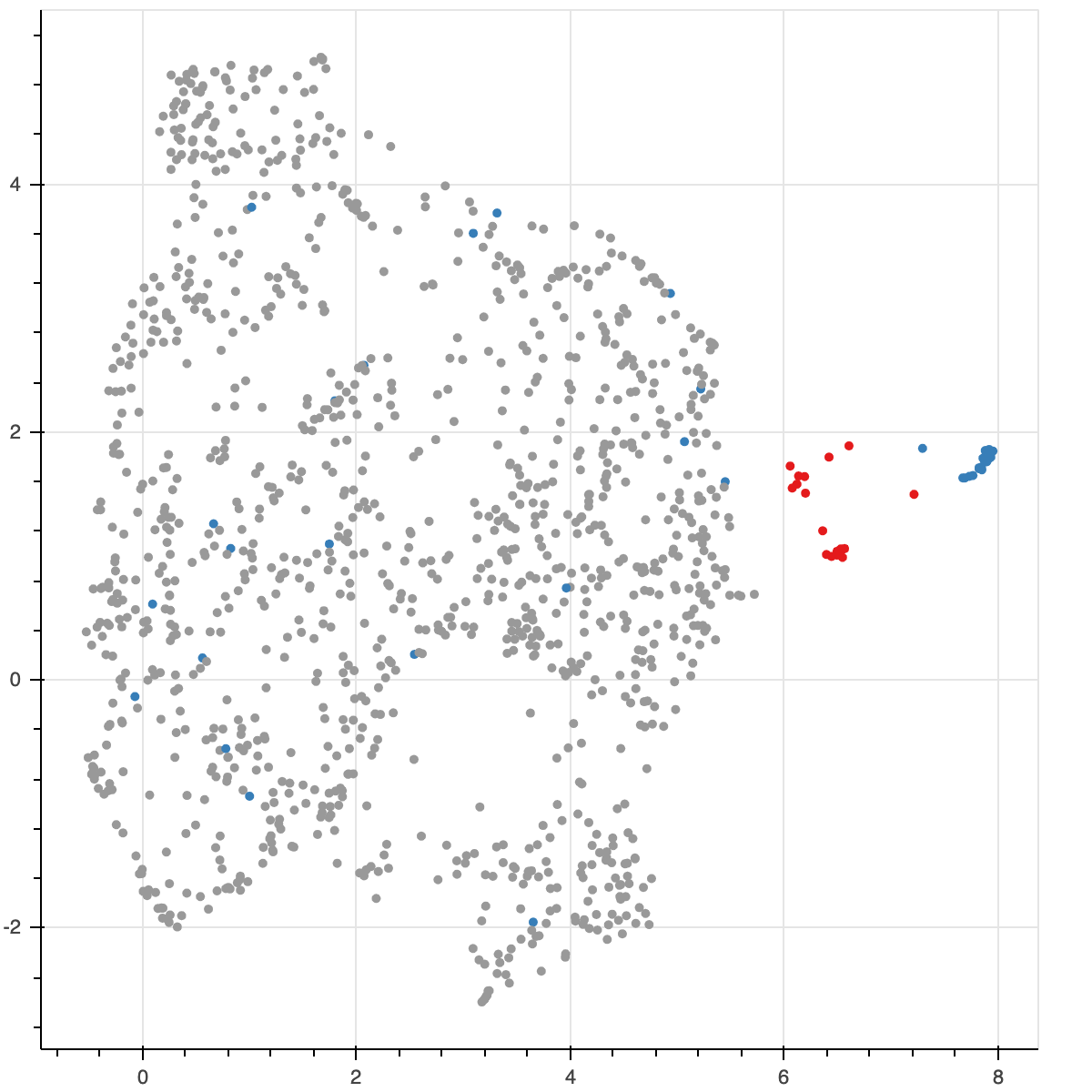}
			\caption{dominant hand arm}
			\label{fig:umap_upper_body}
		\end{subfigure}
		\hfill
		\begin{subfigure}[b]{0.3\textwidth}
			\centering
			\includegraphics[width=0.85\linewidth]{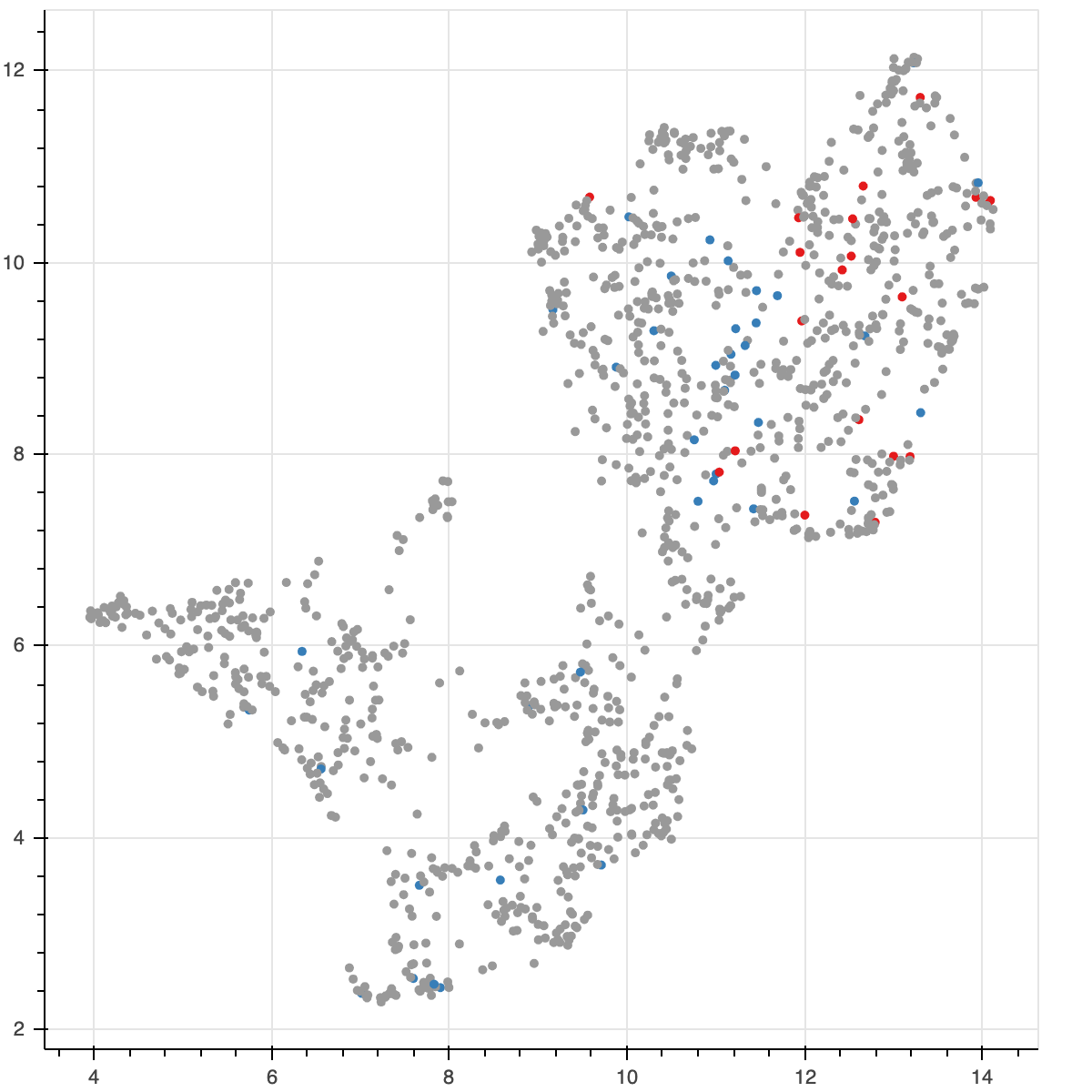}
			\caption{dominant hand wrist}
			\label{fig:umap_upper_body}
		\end{subfigure}
		\caption{UMAP visualizations for one participant for the different skeletal conditions. With red are the signs of the participant and with blue the targeted signs.}
		\label{fig:umpa_vis}
	\end{figure*}

	Highest accuracy is apparent at a top-50 level at 95\% using the UMAP algorithm and the joints of the dominant hand arm. Furthermore, top-20 rank shows an adequate accuracy at 80\% again using UMAP and the dominant hand arm coordinates. Figure \ref{fig:umpa_vis} presents the visualizations of the UMAP algorithm for each of the skeletal condition for one participant. The results of the calculated euclidean distances on the original high-dimensional space show an adequate accuracy of approximately 68\% at the top-50 rank in both dominant hand arm and wrist data-sets. 
	
	Principal component analysis (PCA) performed, on average, better using the wrist coordinates and showed the highest accuracy at the top-50 at approximately 41\%.
	
	DTW showed the highest accuracy at 79\% at top-50 rank using the data of the dominant hand wrist and 77\% using the dominant hand arm. On average, DTW had the best accuracy at around 70\% at the top-20 retrieved signs regardless of the skeletal condition used, with a slight increase noticed using the dominant hand arm data.
	
\begin{figure*}
		\centering
		\begin{subfigure}[b]{0.45\textwidth}
			\centering
			\includegraphics[width=0.9\linewidth]{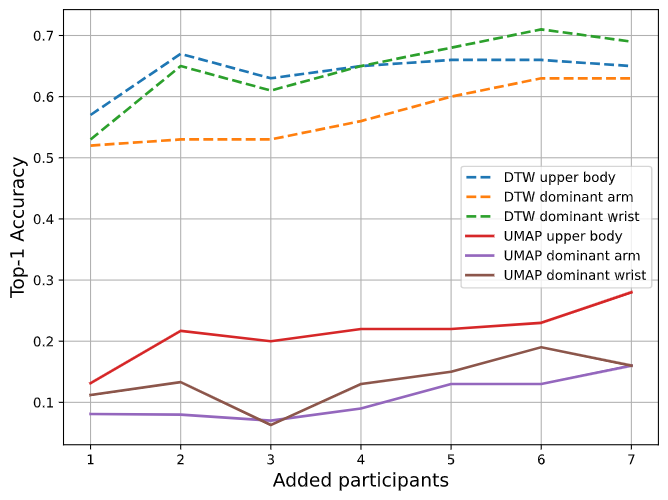}
			\caption{Top-1}
			\label{fig:multiple-top-1}
		\end{subfigure}
		\hfill
		\begin{subfigure}[b]{0.45\textwidth}
			\centering
			\includegraphics[width=0.9\linewidth]{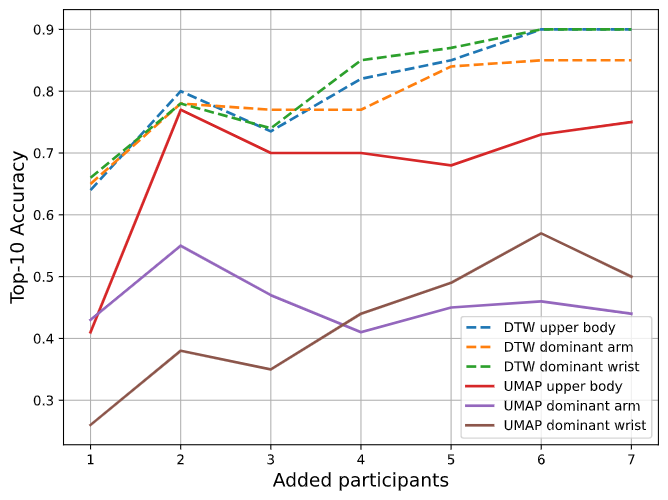}
			\caption{Top-10}
			\label{fig:multiple-yop-10}
		\end{subfigure}
		\caption{Top-1 (a) and Top-10 (b) accuracy using DTW and UMAP based on added participants' data in the lexicon.}
		\label{fig:dtw_more}
	\end{figure*}

	Figure \ref{fig:dtw_more} presents the top-1 and top-10 accuracy levels using DTW and UMAP that have been computed by incrementally adding other participants' data in the lexicon. It can be observed that by adding more sign instances from 6 different participants, the accuracy reached a 90\% level at the top-10 retrieved signs using DTW and the upper body and dominant hand wrist data. Furthermore, a raise of approximately 15\% can be noticed at the top-1 rank on DTW using the upper body and dominant hand wrist joints by adding the data of just 2 participants (Figure 3a). On the other hand, UMAP did not show any adequate raise at the top-1 accuracy regardless of the added participants' data and skeletal condition. However, an increase, of approximately 35\%, can be seen at the top-10 ranking level using the data of 2 participants (Figure 3b).

	\begin{table*}
		\begin{subtable}{1\textwidth}
			\centering
			\resizebox{\textwidth}{!}{\begin{tabular}{|l|c|c|c|c|c|c|c|c|c|c|c|c|}
					\hline
					Skeletal condition & \multicolumn{4}{c|}{Upper body}                                 & \multicolumn{4}{c|}{Dominant hand arm}                         & \multicolumn{4}{c|}{Dominant hand wrist}                              \\ \hline
					Top-k              & Top - 1       & Top - 10      & Top - 20      & Top - 50        & Top - 1         & Top - 10      & Top - 20     & Top - 50      & Top - 1         & Top - 10        & Top - 20        & Top - 50        \\ \hline
					PCA                & 0.0375        & 0.1562        & 0.2187        & 0.325           & 0.025           & 0.1562        & 0.2125       & 0.3437        & 0.0187          & 0.1562          & 0.2125          & 0.4125          \\ \hline
					UMAP               & 0.1312        & 0.4125        & 0.6562        & \textbf{0.6937} & 0.0812          & 0.4375        & \textbf{0.8} & \textbf{0.95} & 0.1125          & 0.2562          & 0.3687          & 0.4575          \\ \hline
					DTW                & \textbf{0.57} & \textbf{0.64} & \textbf{0.67} & 0.687           & \textbf{0.5188} & \textbf{0.65} & 0.7188       & 0.7763        & \textbf{0.5265} & \textbf{0.6562} & \textbf{0.7125} & \textbf{0.7937} \\ \hline
					Euclidean distance & 0.2125        & 0.4625        & 0.5325        & 0.6188          & 0.2625          & 0.5           & 0.6063       & 0.6938        & 0.1875          & 0.445           & 0.55            & 0.675           \\ \hline
			\end{tabular}}
		\end{subtable}
		\label{Tab:full_table}
		\caption{Sign retrieval accuracy per algorithm (by row) on the three skeletal conditions based on the top-k retrieved signs (highest value per column in \textbf{bold}).}
	\end{table*}
	
	\section{Discussion}
	
	In this study we have investigated the use of OpenPose and four different implementations as distance metrics for an efficient ranking pipeline to retrieve matching signs from a sign language lexicon. The results demonstrated that, on a large vocabulary of 1200 signs, such a task can be achieved with an adequate accuracy rate using the Uniform Manifold Approximation and Projection for Dimension Reduction (UMAP) or Dynamic Time Warping (DTW) and the dominant arm joints’ coordinates.
	
	With regard to the visualizations produced by the UMAP algorithm, a few observations can be made. Firstly, by using the upper body joints UMAP produces discrete clusters. These clusters seem to reflect an abstract representation of the movement of each sign. This behavior has been observed also using the dominant hand arm coordinates, although it is more noticeable using all the upper-body joints. We have observed that signs that have similar movement but different handshapes are grouped close to each other. 
	
	However, special consideration needs to be made when viewing the visualizations produced by UMAP, especially the one using the upper body joints. The distances between the noticeable clusters, as well as their size relative to each other, do not hold any particular meaning. This is because of the use of local distances by the algorithm when constructing the graph. However, our results using the euclidean distances on the high-dimensional space suggest that UMAP preserves the original global distances.
	
	Finally, it is worth mentioning that DTW performs equally well irrespective of the skeletal condition used at around 70\% at the top-20 rank. Overall, it produces the most stable and consistent accuracy at the top-10 retrieved signs at around 65\%. This accuracy level can be further raised reaching 90\% by adding 6 more sign instances (from different signers) into the original lexicon. This attribute can be further explored by lexicographers by asking users of their lexica to submit their own versions of signs. This process can significantly boost the performance of DTW in its ability of retrieving the closest matching sign. A broad benefit of using such an algorithm is the fact that lexica compilers do not need to re-train any model if more signs or sign instances are added to their lexica.
	
	In general, while our accuracy does not reach the ones reported by   \cite{schneider_gesture_2019} and \cite{fragkiadakis_signing_2020} (77.4\% top-1 and 74\% top-5 accuracy respectively) using similar algorithms and frameworks, our methods have been applied on a far larger lexicon (1200 signs instead of 6 and 100 respectively). As a result, we provide a better approximation on how these methods can actually be used in real world lexica.
	
	\section{Conclusions}
	To sum up, we have obtained satisfactory results demonstrating that UMAP and DTW, in combination with the pre-trained pose estimation framework OpenPose, can be used as an efficient sign ranking and retrieval system. Our method can effectively be applied to any sign language lexicon without any training process involved. 
	
	To the best of our knowledge, this is the first study using UMAP as a dimensionality reduction technique within the sign language domain and showcasing the strength of such algorithm compared to other implementations. 
	
	Future work will focus on exploring additional deep learning implementations for an efficient handshape and pose recognition. Their use, as well as supplementary hyper-parameter optimization for the techniques used in this study, could lead to an increase in accuracy. Whilst as we argued that for the availability of dictionaries it will be good to focus on zero training approaches, in future work we intend to run comparative analysis to understand the impact of training based approaches on performance.
	We propose that further research should also be undertaken in order to assess the application of our method on different datasets and languages.  On a wider level, the techniques used in this study could be further explored to measure variation in different sign languages. The results from the use of UMAP and DTW on a large vocabulary suggest that these techniques might be well suited for variation measurement tasks, broadening their use beyond the search functionality for sign language lexica. 	
	
	\small
	
	\bibliographystyle{apalike}
	\bibliography{mtsummit2021}

\end{document}